\documentclass[journal,transmag]{IEEEtran}
\ifCLASSINFOpdf
\else
\fi
\usepackage{amsmath}
\usepackage{graphicx}
\usepackage{subcaption}
\usepackage{hyperref}
\usepackage{multirow}
\begin{document}
%
\title{UFLUX v2.0: A Process-Informed Machine Learning Framework for Efficient and Explainable Modelling of Terrestrial Carbon Uptake}


\author{Wenquan Dong, Songyan Zhu, Jian Xu, \textit{Senior Member, IEEE},
        Casey M. Ryan, Man Chen, Jingya Zeng,\\ Hao Yu, \textit{Student Member, IEEE}, Congfeng Cao, Jiancheng Shi, \textit{Fellow, IEEE}\\
\thanks{Manuscript received XX XX 2024; (Corresponding author: Songyan Zhu).} 
\thanks{\fontfamily{ptm}\selectfont Wenquan Dong is with the School of Geosciences, University of Edinburgh, Edinburgh EH9 3FF, UK, and also with Royal Botanic Garden Edinburgh, Edinburgh EH3 5LR, UK.} 
\thanks{\fontfamily{ptm}\selectfont Songyan Zhu is with the Faculty of Environmental and Life Sciences, University of Southampton, Southampton SO17 1BJ, UK (e-mail: songyan.zhu@soton.ac.uk).}
\thanks{\fontfamily{ptm}\selectfont Jian Xu and Jiancheng Shi are with the National Space Science Center, Chinese Academy of Sciences, Beijing 100190, China}
\thanks{\fontfamily{ptm}\selectfont Casey M. Ryan is with the School of Geosciences, University of Edinburgh, Edinburgh EH9 3FF, UK.} 
\thanks{\fontfamily{ptm}\selectfont Man Chen is with LERMA/Observatoire de Paris, Paris 75014, France, and also with Sorbonne University, Paris 75005, France.} 
\thanks{\fontfamily{ptm}\selectfont Jingya Zeng is with the Department of Economics, Faculty of Environment, Science and Economy, University of Exeter, EX4 4PU Exeter, U.K.}
\thanks{\fontfamily{ptm}\selectfont Hao Yu is with the School of Engineering, Institute for Imaging, Data and Communications, University of Edinburgh, Edinburgh EH9 3JL, UK.}
\thanks{\fontfamily{ptm}\selectfont Congfeng Cao is with the Institute for Logic, Language and Computation, University of Amsterdam, Amsterdam 1098 XH, Netherlands.}

}

\markboth{}
{Shell \MakeLowercase{\textit{et al.}}: A Sample Article Using IEEEtran.cls for IEEE Journals}
%



\maketitle

\begin{abstract}
Gross Primary Productivity (GPP), the amount of carbon plants fixed by photosynthesis, is pivotal for understanding the global carbon cycle and ecosystem functioning. Process-based model built on the knowledge of ecological processes, are susceptible to biases stemming from their assumptions and approximations. These limitations potentially result in considerable uncertainties in global GPP estimation, which may pose significant challenges to our Net Zero goals. This study presents UFLUX v2.0, a process-informed model that integrates state-of-art ecological knowledge and advanced machine learning technique to reduce uncertainties in GPP estimation by learning the biases between process-based models and eddy covariance (EC) measurements. In our findings, UFLUX v2.0 demonstrated a substantial improvement in model accuracy, achieving an R\textsuperscript{2} of 0.79 with a reduced RMSE of 1.60\,{g\,C\,m$^{-2}$\,d$^{-1}$}, compared to the process-based model’s R\textsuperscript{2} of 0.51 and RMSE of 3.09\,{g\,C\,m$^{-2}$\,d$^{-1}$}. Our global GPP distribution analysis indicates that while UFLUX v2.0 and the process-based model achieved similar global total GPP (137.47\,{Pg\,C} and 132.23\,{Pg\,C}, respectively), they exhibited large differences in spatial distribution, particularly in latitudinal gradients. These differences are very likely due to systematic biases in the process-based model and differing sensitivities to climate and environmental conditions. This study offers improved adaptability for GPP modelling across diverse ecosystems, and further enhances our understanding of global carbon cycles and its responses to environmental changes. 

\end{abstract}




%
\IEEEpeerreviewmaketitle

\section{Introduction}
%
%
%
%
\IEEEPARstart{T}{errestrial} ecosystems play a crucial role in the global carbon cycle, with GPP serving as a key metric for quantifying the carbon uptake by vegetation through photosynthesis \cite{beer2010terrestrial}. Accurate estimation of GPP is fundamental for understanding ecosystem functioning, assessing carbon budgets, and thus mitigating climate change \cite{piao2013evaluation, chen2021seasonal}. However, the spatial and temporal heterogeneity of terrestrial ecosystems poses significant challenges to precise GPP quantification at regional and global scales.

The EC measurements have emerged as a valuable solution for directly measuring carbon fluxes at the ecosystem level, offering continuous, high-accuracy data that are crucial for understanding ecosystem-atmosphere interactions and the global carbon cycle \cite{aubinet2012eddy}. Despite their importance, EC measurements are limited in spatial coverage due to the high cost of maintaining flux towers and technical requirements \cite{zhu2024uflux}. These limitations highlight the necessity for robust models that can extend GPP estimation across diverse landscapes and climatic conditions where EC data are unavailable.

The application of process-based models, such as the light-use efficiency models, has gained traction for estimating GPP over large scales \cite{yuan2007deriving}. These models leverage our understanding of physiological and ecological processes to simulate carbon fluxes based on environmental drivers and vegetation characteristics. The adoption of the optimality principle particularly offers great advantages in predicting light use efficiency (LUE) without the need for predefined vegetation-type-specific parameters; this approach provides a more generalised method for GPP estimation across diverse biomes \cite{stocker2020p}. This approach allows for broad applicability, making it a valuable tool for global-scale simulations. Moreover, process-based models have been rigorously evaluated against GPP derived from EC flux measurements (using the FLUXNET 2015 Tier 1 dataset), demonstrating its relevance in aligning model predictions with site-specific observations. However, process-based models often struggle to fully capture the complex interactions within ecosystems, potentially leading to biases in GPP estimates \cite{rogers2017roadmap}. 

These biases in process-based modelling can arise from simplifications in model structure, uncertainties in parameter values, or limitations in representing the full spectrum of ecosystem processes. While the LUE model addresses some of these challenges by incorporating factors such as low temperature effects on quantum yield and soil moisture stress, it still faces inherent limitations in accurately simulating the diverse range of ecosystem responses to environmental variables \cite{stocker2020p}. These limitations underscore the need for continuous refinement and validation of process-based models to improve their accuracy and applicability across different ecosystems and climatic conditions.

In recent years, machine learning (ML) approaches have gained popularity as powerful tools for improving GPP estimations \cite{sarkar2022machine, sun2019evaluating}. These data-driven methods can capture intricate relationships between environmental variables and GPP, potentially overcoming some limitations of process-based models. ML models have demonstrated superior performance in various ecological applications \cite{dong2024new, zhu2023explainable}. However, despite their predictive power, ML models often lack the mechanistic interpretability of process-based models and may struggle to extrapolate accurately beyond the conditions represented in their training data, potentially limiting their reliability in scenarios of environmental change or in data-scarce regions \cite{wolanin2019estimating}.

Building upon the success of UFLUX \cite{zhu2024uniform}, we present UFLUX v2.0, an advanced machine learning framework designed to enhance GPP estimation accuracy. UFLUX v2.0 innovatively learns the biases between process-based model estimations and site measurements, aiming to make these biases ecologically explainable. This approach represents a novel integration of process-based understanding and data-driven techniques, potentially offering a more robust and accurate method for GPP estimation across diverse ecosystems.

\section{Methodology}
\subsection{UFLUX v2.0}
The UFLUX v2.0 framework represents a comprehensive approach to estimating GPP. This framework is designed to address the inherent limitations of traditional process-based models by incorporating an adaptive bias correction mechanism, thereby improving the accuracy and generalizability of GPP estimates across diverse ecosystems.

\begin{figure} [htbp]
    \centering    
        \centering
        \includegraphics[width=\columnwidth]{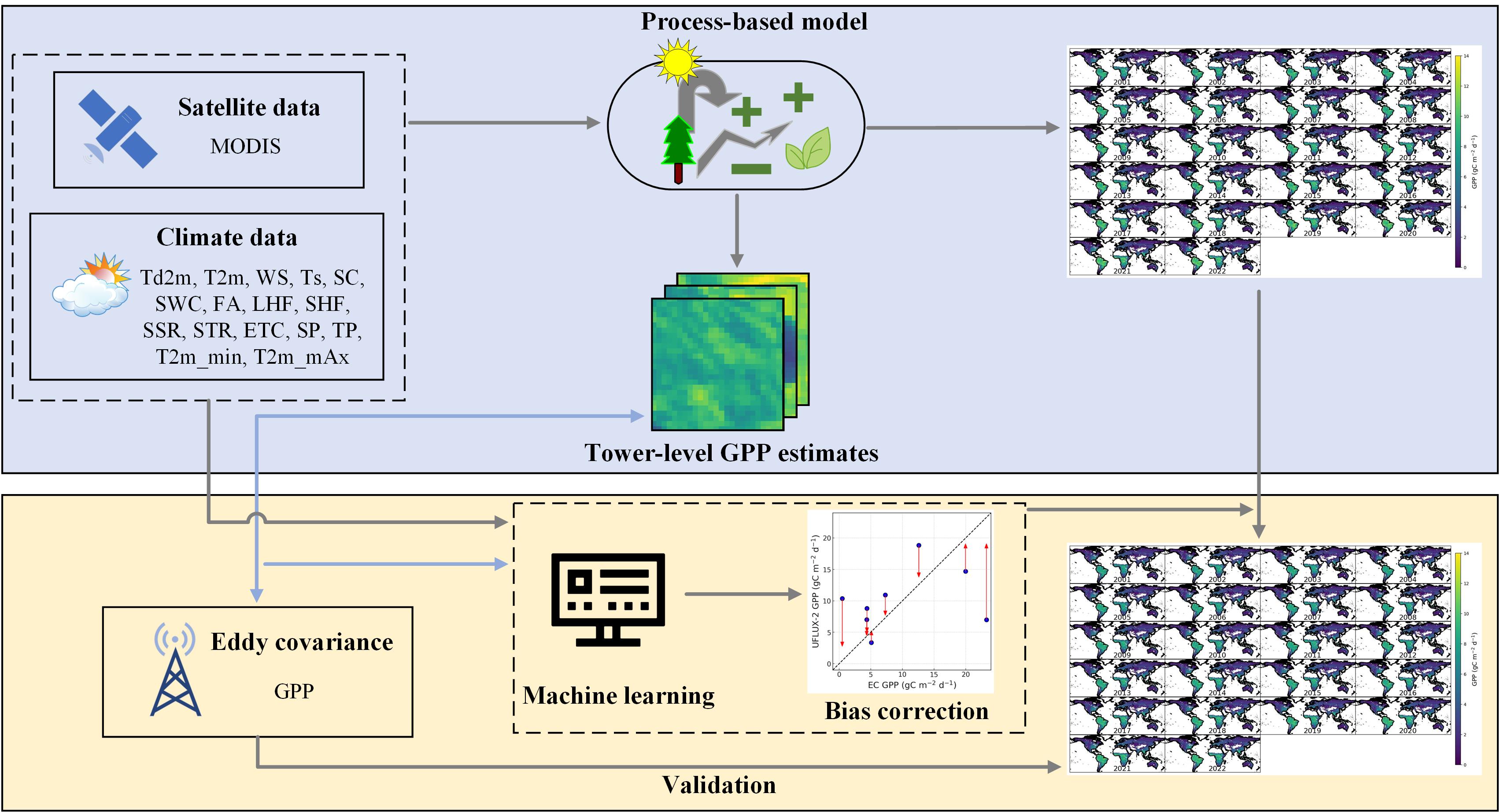}  

\caption{Schematic workflow of the UFLUX v2.0 framework for GPP estimation. The upper panel shows the process-based model component, integrating satellite and climate data. The lower panel illustrates the UFLUX v2.0 enhancement, incorporating machine learning for adaptive bias correction using eddy covariance measurements, resulting in improved global GPP estimates.}
\label{fig: workflow}
\end{figure}

The framework begins by fetching satellite observations and climate/environmental data, which provide crucial inputs for characterizing vegetation dynamics and environmental conditions (Fig.~\ref{fig: workflow}). These data are then used by a selected process-based model, as UFLUX v2.0 is designed to be flexible in its choice of process-based models, allowing for the integration of the most suitable option for a given context. Process-based models offer the advantage of simulating photosynthetic processes based on established ecological principles. However, despite their strengths, these models often struggle with biases and inaccuracies due to their inherent assumptions and simplifications. Recognizing the potential biases and limitations of purely process-based models, UFLUX v2.0 incorporates an innovative adaptive bias correction module. This bias correction is implemented using a machine learning algorithm, which learns the relationship between the discrepancies of the process-based model predictions and observed GPP from EC measurements. By training on these discrepancies, the bias correction model dynamically adjusts the GPP predictions to improve accuracy. In this study, we employed XGBoost, a highly efficient gradient boosting algorithm, though UFLUX v2.0 can accommodate various machine learning methods depending on the specific requirements of the application or the nature of the data.

For model validation, we used EC data to conduct a 5-fold cross-validation, assessing the model's performance across different subsets of the sites. The global GPP map presented is the average of the five cross-validation folds, providing a robust representation of GPP while accounting for potential variability in model predictions.

\subsection{Process-based model} 
The process-based model used in this study is the LUE-based model \cite{stocker2020p}. The LUE-based model integrates the Farquhar-von Caemmerer-Berry (FvCB) model for C$_3$ photosynthesis with an optimality principle for the carbon assimilation-transpiration trade-off \cite{stocker2020p}. It predicts GPP as:

\begin{equation}
\text{GPP} = \text{PAR} \cdot \text{fAPAR} \cdot \text{LUE}
\end{equation}

where PAR is incident photosynthetically active radiation, and fAPAR is the fraction of PAR absorbed by green tissue.
The key innovation of the P-model is its prediction of LUE based on first principles. It assumes an optimal ratio of leaf-internal to ambient CO$_2$ concentration ($\chi = c_i/c_a$) that balances the costs of maintaining transpiration and carboxylation capacities:

\begin{equation}
\chi = \frac{\Gamma^*}{c_a} + \left(1 - \frac{\Gamma^*}{c_a}\right) \frac{\xi}{\xi + \sqrt{D}}
\end{equation}

where $\Gamma^*$ is the photorespiratory compensation point, c$_a$ is ambient CO$_2$ concentration, and $\xi$ is a parameter related to the carbon-water trade-off, while D is vapor pressure deficit \cite{wang2017towards}. Specifically, $\xi$ is defined as:

\begin{equation}
\xi = \sqrt{\frac{\beta(K + \Gamma^*)}{1.6\eta^*}}
\end{equation}

where $\beta$ is the ratio of unit costs for carboxylation to transpiration, K is the effective Michaelis-Menten coefficient for RuBisCO-limited assimilation, and $\eta^*$ is the viscosity of water relative to its value at 25°C. This relationship ensures that the model dynamically adjusts to environmental conditions, providing more accurate predictions of GPP under varying climates.

The model further assumes a coordination hypothesis, which posits that the photosynthetic machinery operates near the intersection of light-limited and RuBisCO-limited assimilation rates. This coordination is captured by a relationship between the maximum rate of RuBisCO carboxylation (V${c\text{max}}$) and the maximum rate of electron transport (J${\text{max}}$). This leads to a formulation of LUE as:

\begin{equation}
\text{LUE} \mathrel{\widehat{=}} \varphi_0(T) \, \beta(\theta) \, m' \, M_C
\end{equation}

where $\phi_0(T)$ is the temperature-dependent intrinsic quantum yield of photosynthesis, $\beta(\theta)$ is a soil moisture stress factor, $m'$ is a factor accounting for the co-limitation of light and RuBisCO, and M$_C$ is the molar mass of carbon.

The LUE-based model thus provides a mechanistic basis for predicting LUE across diverse environmental conditions, thereby reducing the need for prescribed vegetation-type-specific parameters. It applies a unified framework to different ecosystems by predicting photosynthetic parameters rather than relying on fixed values. The model has been shown to perform well in predicting GPP across a global network of flux measurement sites, offering a powerful tool for large-scale ecosystem modeling.

\subsection{Data}
We used three types of data, including GPP data from 206 EC towers, vegetation index derived from satellite data, and environmental drivers.

\subsubsection{Eddy Covariance}
This study uses the FLUXNET2015 dataset, which consists high-frequency eddy covariance measurements from a global network of flux towers \cite{pastorello2020fluxnet2015}. The FLUXNET2015 database provides data from 206 open-access towers representing a wide range of biomes and climatic conditions. The dataset employs standardized data processing techniques, including advanced gap-filling methods and data quality assessments, ensuring consistency and reliability of measurements across sites.

For our analysis, we primarily used GPP estimates derived from net ecosystem exchange measurements from the FLUXNET2015 dataset. The eddy covariance towers were classified into 12 different plant functional types (PFTs): croplands (CROs), closed shrublands (CSHs), deciduous broadleaf forests (DBFs), deciduous needleleaf forest (DNF), evergreen broadleaf forests
(EBFs), evergreen needleleaf forests (ENFs), grasslands (GRA), mixed forests (MF), open shrublands (OSHs), savannahs (SAVs), permanent wetlands (WETs), and woody savannahs (WSAs) \cite{pastorello2020fluxnet2015}.

\subsubsection{Remote sensing data}
This study incorporates remote sensing data from the MCD43A4 product, which provides surface reflectance at a 500-meter spatial resolution \cite{schaaf2002first}. We used the red (620–670 nm) and near-infrared (841–876 nm) spectral bands due to their relevance in vegetation monitoring. From these spectral bands, we calculated the Near-Infrared Reflectance of Vegetation (NIRv), a key metric that integrates near-infrared reflectance and the normalized difference vegetation index (NDVI), which has been shown to be strongly correlated with photosynthetic activity and GPP \cite{badgley2019terrestrial}. The NIRv is computed using the following formula:

\begin{equation} 
\text{NIRv} = \text{NDVI} \times \text{NIR}_{ref} 
\end{equation}

where NDVI is defined as:

\begin{equation} 
\text{NDVI} = \frac{\text{NIR}_{ref} - \text{Red}_{ref}}{\text{NIR}_{ref} + \text{Red}_{ref}} 
\end{equation}

where $\text{NIR}_{ref}$ represents the reflectance in the near-infrared band and $\text{Red}_{ref}$ represents the reflectance in the red band.

\subsubsection{climate reanalysis}
For the climate variables, we used the ERA5 Daily Aggregates dataset, produced by the European Centre for Medium-Range Weather Forecasts (ECMWF). ERA5 is the fifth generation ECMWF atmospheric reanalysis of the global climate, providing a comprehensive record of the global atmosphere, land surface, and ocean waves from 1950 onwards \cite{hersbach2020era5}. From the ERA5 Daily Aggregates, we obtained a comprehensive set of climate variables including 2 m air temperature, dewpoint temperature, soil temperature at surface level, snow cover, volumetric soil water content in the top soil layer, forecast albedo, surface latent heat flux, surface sensible heat flux, downward surface solar radiation, downward surface thermal radiation, evaporation from the top of the canopy, surface pressure, total precipitation, minimum and maximum daily 2 m air temperatures, and the wind components at 10 m (u and v). Additionally, vapour pressure deficit (VPD) was calculated using the air temperature and dewpoint temperature.

\section{Results and Discussion}
\subsection{Tower-Level Validation}
\begin{figure} [htbp]
    \centering    
        \centering
        \includegraphics[width=\columnwidth]{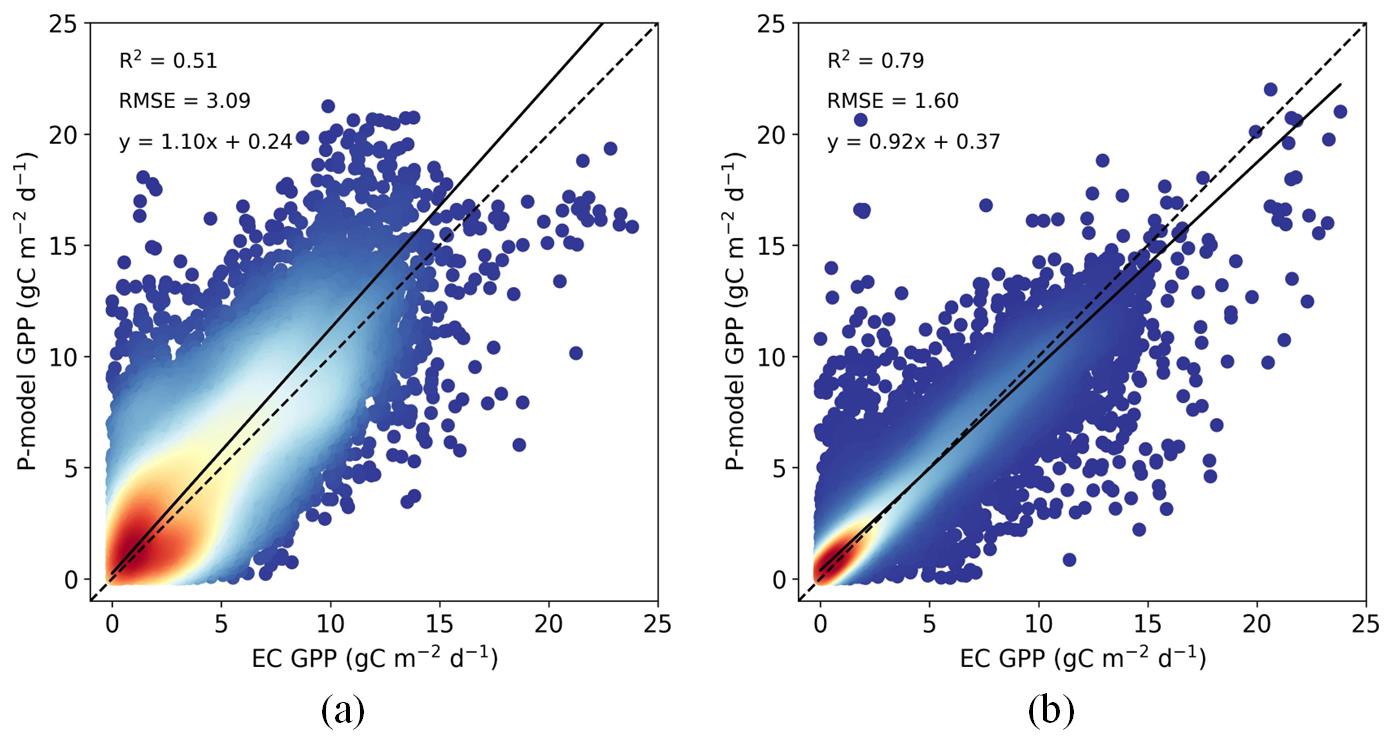}  

\caption{Comparison of modeled GPP estimates against EC GPP measurements at the tower level. (a) Process-based model GPP estimates versus EC GPP, and (b) UFLUX v2.0 GPP estimates versus EC GPP. The dashed line represents the 1:1 line. The color of the scatter points indicates the point density, calculated using a Gaussian kernel density estimation, with colors ranging from dark blue to yellow. Darker colors represent areas with lower data density, while lighter colors (yellow) indicate areas of higher data density.}
\label{fig: EC_validation}
\end{figure}

The performance of the UFLUX v2.0 model in estimating GPP was evaluated against the process-based model using EC measurements. As shown in Fig. \ref{fig: EC_validation}, the process-based model achieves an R\textsuperscript{2} of 0.51 with an RMSE of 3.09\,{g\,C\,m$^{-2}$\,d$^{-1}$}, indicating moderate accuracy but with notable scatter. In contrast, UFLUX v2.0 demonstrates a marked improvement in performance, achieving an R\textsuperscript{2} of 0.79 and reducing the RMSE to 1.60\,{g\,C\,m$^{-2}$\,d$^{-1}$}.

\begin{table}[htbp]
\centering
\caption{COMPARISON OF PROCESS-BASED MODEL AND UFLUX V2.0 PERFORMANCE METRICS}
\label{tab:model-comparison}
\begin{tabular}{lccccccc}
\hline
\multirow{2}{*}{PFTs} & \multirow{2}{*}{Measurements} & \multicolumn{2}{c}{Process-based model} & \multicolumn{2}{c}{UFLUX v2.0} \\
\cline{3-6}
 &  & R$^2$ & RMSE & R$^2$ & RMSE \\
\hline
CRO & 287 & 0.57 & 3.59 & 0.68 & 2.98 \\
CSH & 37  & 0.48 & 1.92 & 0.82 & 0.71 \\
DBF & 315 & 0.79 & 3.03 & 0.89 & 1.50 \\
DNF & 7   & 0.73 & 1.82 & 0.37 & 0.69 \\
EBF & 166 & 0.46 & 2.65 & 0.82 & 1.26 \\
ENF & 600 & 0.69 & 2.57 & 0.89 & 1.04 \\
GRA & 469 & 0.64 & 2.91 & 0.87 & 1.24 \\
MF  & 187 & 0.67 & 2.70 & 0.91 & 1.05 \\
OSH & 101 & 0.52 & 2.31 & 0.63 & 1.00 \\
SAV & 104 & 0.45 & 3.40 & 0.63 & 1.44 \\
WET & 149 & 0.64 & 2.64 & 0.71 & 1.80 \\
WSA & 144 & 0.53 & 2.74 & 0.86 & 0.94 \\
\hline
\end{tabular}
\end{table}

Although the process-based model's performance varied across the 12 PFTs, it generally achieved reasonably good results (see Table \ref{tab:model-comparison}). However, UFLUX v2.0 consistently demonstrated stronger correlations between modeled and observed GPP, as evidenced by higher R\textsuperscript{2} values and lower RMSE across most PFTs (see Table \ref{tab:model-comparison}). In forest ecosystems, UFLUX v2.0 showed marked improvements. For ENF, the R\textsuperscript{2} increased from 0.69 to 0.89, while RMSE decreased from 2.57\,{g\,C\,m$^{-2}$\,d$^{-1}$} to 1.04\,{g\,C\,m$^{-2}$\,d$^{-1}$}. Similar enhancements were observed in DBF and evergreen EBF, suggesting UFLUX v2.0's superior capability in capturing complex forest ecosystem dynamics. GRA and CSH also saw substantial improvements, with GRA's R\textsuperscript{2} increasing from 0.64 to 0.87 and RMSE decreasing from 2.91\,{g\,C\,m$^{-2}$\,d$^{-1}$} to 1.24\,{g\,C\,m$^{-2}$\,d$^{-1}$}.

Despite overall improvements, both models faced challenges in certain ecosystems. CRO proved difficult, likely due to the complexity of agricultural management practices, though UFLUX v2.0 showed slight improvement with R\textsuperscript{2} increasing from 0.57 to 0.68, and RMSE decreasing from 3.59\,{g\,C\,m$^{-2}$\,d$^{-1}$} to 2.98\,{g\,C\,m$^{-2}$\,d$^{-1}$}. Additionally, while UFLUX v2.0 improved estimates for OSH and SAV, the overall performance in these categories remained relatively weak, suggesting a need for further refinement in modeling arid or semi-arid ecosystems. It is noteworthy that UFLUX v2.0 struggled with DNF, possibly due to limited sample size of this PFT. 

The enhanced performance of UFLUX v2.0 across diverse PFTs underscores the importance of integrating data-driven techniques with process-based models. The inherent limitations of the process-based model, which are largely due to its generalized assumptions about ecosystem processes, are effectively mitigated by the adaptive bias correction in UFLUX v2.0. UFLUX v2.0 not only improves the precision of GPP estimates but also extends the model's applicability across different ecosystems, including those with complex environmental interactions that are not easily captured by process-based models.

\subsection{Global distribution of GPP}
\begin{figure}[htbp]
    \centering    
        \centering
        \includegraphics[width=\columnwidth]{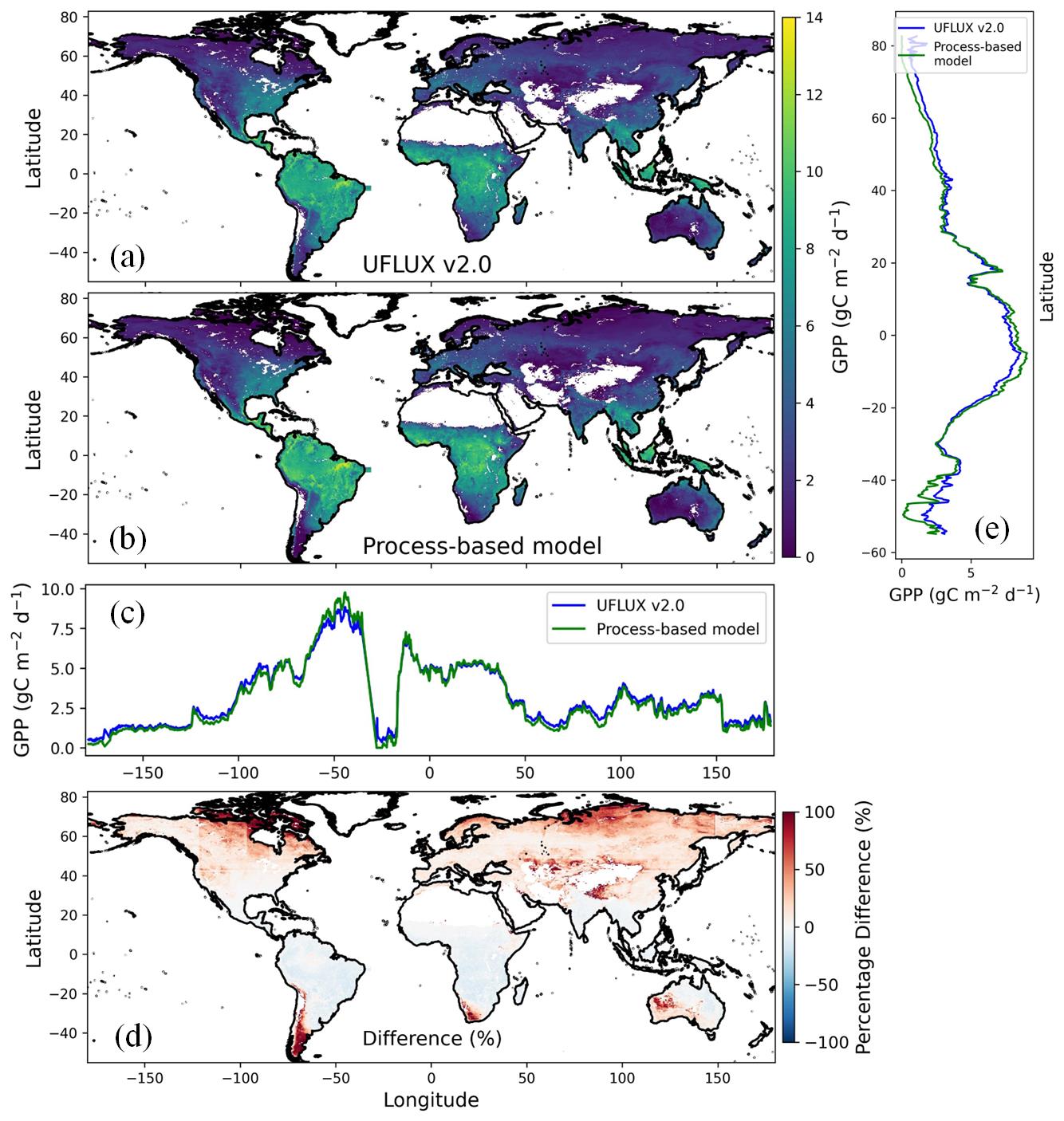}  

\caption{Global distribution and comparison of GPP estimates from UFLUX v2.0 and the process-based model for the year 2010. (a) Global GPP estimated by UFLUX v2.0. (b) Global GPP estimated by the process-based model. (c) Longitudinal distribution of GPP for both models. (d) Percentage difference in GPP estimates between UFLUX v2.0 and the process-based model, calculated as (UFLUX v2.0 - Process-based model) / UFLUX v2.0 × 100\%. (e) Latitudinal distribution of GPP for both models.}
\label{fig: GPP_distribution}
\end{figure}

To compare the global distribution of GPP between UFLUX v2.0 and the process-based model, we used the GPP estimates for the year 2010 as a case study. UFLUX v2.0 predicts a total GPP of 137.47 Pg C, while the process-based model estimates a slightly lower value of 132.23 Pg C. Fig. \ref{fig: GPP_distribution} illustrates the spatial and latitudinal-longitudinal distribution of GPP for both models.

Both models capture similar broad-scale patterns in GPP distribution, with peak values in tropical regions and decreasing productivity towards higher latitudes. However, notable differences emerge in the spatial comparison [Fig. \ref{fig: GPP_distribution} (a), (b), (d)]. UFLUX v2.0 generally yields lower GPP values in tropical and temperate regions compared to the process-based model, particularly evident in high-productivity areas such as the Amazon, central Africa, and Southeast Asia. This moderation of values by UFLUX v2.0 suggests that its adaptive bias correction mechanism effectively mitigates potential overestimation tendencies of the process-based model.

The latitudinal profile [Fig. \ref{fig: GPP_distribution} (c)] further emphasizes the differences between the models. Both UFLUX v2.0 and the process-based model show peak GPP values near the equator, with a gradual decline towards higher latitudes. Notably, the process-based model estimates slightly higher GPP values in the tropics (between 20°S and 20°N) compared to UFLUX v2.0. In contrast, UFLUX v2.0 tends to estimate marginally higher GPP values in the mid to high latitudes (north of 40°N and south of 40°S). The pronounced latitudinal differences may be attributed to the models' varying sensitivities to factors such as temperature gradients, day length variations, and distinct vegetation adaptations across latitudinal bands. Both models show similar patterns of GPP distribution along longitude [Fig. \ref{fig: GPP_distribution} (e)]. They estimate higher GPP values between 40°W and 80°W, and between 10°W and 50°E, covering areas such as the Amazon and Central Africa.

\section{Summary and Conclusion}
In this study, we introduced UFLUX v2.0, an advanced framework for estimating GPP that integrates process-based modeling with machine learning techniques. Validating against data from 206 FLUXNET2015 sites, UFLUX v2.0 demonstrated enhanced performance across various plant functional types compared to a process-based model.

While the global GPP estimates for 2010 were relatively similar between the two models---137.47\,{Pg\,C} for UFLUX v2.0 and 132.23\,{Pg\,C} for the process-based model---the spatial distribution of GPP revealed substantial differences. These differences were most pronounced across latitudinal gradients, with UFLUX v2.0 producing lower GPP estimates in tropical regions and higher estimates at higher latitudes (north of 40°N and south of 40°S). In contrast, the longitudinal distribution showed less variation between the two models.

In conclusion, UFLUX v2.0 offers a robust solution for enhancing GPP estimates across diverse ecosystems by addressing the inherent biases in process-based models. Its integration of machine learning techniques with ecological modeling holds promise for improving large-scale carbon cycle assessments, with implications for both scientific research and climate change mitigation efforts.


%



\ifCLASSOPTIONcaptionsoff
  \newpage
\fi



\bibliographystyle{IEEEtran}
%

  

\bibliography{reference.bib}

%




\end{document}